\documentclass[sigconf]{acmart}
\AtBeginDocument{%
  \providecommand\BibTeX{{%
    \normalfont B\kern-0.5em{\scshape i\kern-0.25em b}\kern-0.8em\TeX}}}

\usepackage{graphicx}
\usepackage{amsfonts} 
\usepackage{algorithm} 
\usepackage{algpseudocode} 
\usepackage{float}
\copyrightyear{2023}
\acmYear{2023}
\setcopyright{acmlicensed}\acmConference[SOICT 2023]{The 12th International Symposium on Information and Communication Technology}{December 7--8, 2023}{Ho Chi Minh, Vietnam}
\acmBooktitle{The 12th International Symposium on Information and Communication Technology (SOICT 2023), December 7--8, 2023, Ho Chi Minh, Vietnam}
\acmPrice{15.00}
\acmDOI{10.1145/3628797.3628983}
\acmISBN{979-8-4007-0891-6/23/12}

\begin{document}

\title{ALGNet: Attention Light Graph Memory Network for Medical Recommendation System}

\author{Minh-Van Nguyen $^{1,2}$, Duy-Thinh Nguyen $^{1,2}$, Quoc-Huy Trinh$^{1,2}$, and Bac Le$^{1,2}$}
\email{20127094@student.hcmus.edu.vn, 20127333@student.hcmus.edu.vn }
\email{20120013@student.hcmus.edu.vn, lhbac@fit.hcmus.edu.vn}
\orcid{0000-0002-5681-472X}

\affiliation{%
    $^1$\institution{Faculty of Information Technology, University of Science, VNU-HCM}
  \city{Ho Chi Minh City}
  \country{Vietnam}
}
\affiliation{%
   $^2$\institution{Vietnam National University, Ho Chi Minh City, Vietnam}
  \city{Ho Chi Minh City}
  \country{Vietnam}
}
\renewcommand{\shortauthors}{Minh-Van Nguyen, et al.}

\begin{abstract}
Medication recommendation is a vital task for improving patient care and reducing adverse events. However, existing methods often fail to capture the complex and dynamic relationships among patient medical records, drug efficacy and safety, and drug-drug interactions (DDI). In this paper, we propose ALGNet, a novel model that leverages light graph convolutional networks (LGCN) and augmentation memory networks (AMN) to enhance medication recommendation. LGCN can efficiently encode the patient records and the DDI graph into low-dimensional embeddings, while AMN can augment the patient representation with external knowledge from a memory module. We evaluate our model on the MIMIC-III dataset and show that it outperforms several baselines in terms of recommendation accuracy and DDI avoidance. We also conduct an ablation study to analyze the effects of different components of our model. Our results demonstrate that ALGNet can achieve superior performance with less computation and more interpretability. The implementation of this paper can be found at: https://github.com/tivivui95/ALGNet
\end{abstract}




\maketitle

\section{Introduction}
The proliferation of digital platforms and the exponential growth of data have propelled recommendation systems to the forefront of research and application across various domains \cite{das2017survey}. These systems, fueled by sophisticated algorithms and data-driven insights, play a pivotal role in assisting users by providing personalized suggestions, thereby enhancing their decision-making processes \cite{chen2013human}. Among the myriad of fields that benefit from recommendation systems, the healthcare domain stands out as a critical arena where the seamless integration of these systems holds immense potential to transform patient care, medical research, and clinical operations. This thesis endeavors to delve into the intricate interplay between recommendation systems and the healthcare sector, with a special focus on the enhancement of the GAMEnet baseline model using the innovative LGCN \cite{he2020lightgcn} technique.

The healthcare sector, characterized by its vast and intricate network of patient profiles, medical histories, and treatment options, stands to gain substantially from the implementation of recommendation systems. In the context of patient care \cite{sujan2019human}, these systems can offer personalized treatment plans, medication suggestions, and preventive measures based on patient medical records and historical data. Moreover, healthcare professionals can benefit from recommendations for relevant research articles, clinical trials, and diagnostic methodologies, optimizing their decision-making processes.

The advent of GNNs \cite{fan2019graph} has introduced a paradigm shift in recommendation system research. GNNs excel in handling data with complex relationships and dependencies, making them an ideal candidate for scenarios where entities and their interactions can be represented as graphs. In the context of recommendation systems, GNNs can capture latent connections between users, items, and their attributes, leading to more accurate and interpreted recommendations.

On the other side, GAMEnet is one of the most innovative framework that merges a knowledge graph of DDI \cite{rodrigues2019drug} through a memory module utilizing GCN \cite{zhang2019graph}. This amalgamation interfaces with longitudinal patient records, effectively treated as queries. The model is meticulously trained as an end-to-end system, engineered to deliver both secure and individualized recommendations for medication combinations. Remarkably, GAMEnet outperforms all current benchmark methods across a spectrum of effectiveness metrics.

Despite the impressive performance, GAMENet still has several limitations. Firstly, the usage of the Dual RNN and the GCN of the GAMENet can increase the computation cost. Moreover, the RNN components may not allow the model to learn the long-term dependencies due to the vanishing gradient. This can create negative impacts on the medication recommendation because the medical report statistics have correlations over the long period. To overcome these limitations, we propose the ALG Net, an Attention Light Graph Memory Network for Medical Recommendation Systems, which can capture the long-range dependency of the medical data, additionally, we optimize the computation cost of the graph by applying the LGCN for the graph creation in the recommendation task. Our proposed method is the modification of the GAMENet concept which is described in the rest of the paper.

In a broader context, LGCN have the capacity to yield significant economic benefits for businesses by effectively addressing various challenges associated with cost management, particularly within the domain of graph-based modeling approaches. Furthermore, Dual-Self-Attention plays a crucial role in managing and generating valuable insights from intricate structural data within the healthcare industry, ensuring attention to intricate relationships and robust output generation but still lead to a low cost management.

To sum up, we have the following contributions:
\begin{itemize}

    \item We introduce the ALGNet, which is the modification of the GAMENet to address the challenge of the lack of long-range dependency information, and the computation cost optimization.

    \item The Dual-Self-Attention modules are introduced to capture the long-range dependency of the medical records. 

    \item We investigate the replacement of the GCN by the Light-GCN to improve the computation cost of the previous methods, also we explore the fusion to enhance the recommendation result.

    \item The effectiveness of the ALGNet is evaluated with our methods on the real EHR data. Our method achieves the competitive result in State-of-the-art.
    
\end{itemize}

The rest of the paper is organized as follows. The ~\ref{sec:RelatedWork} will review and describe the task we are doing, and review the previous methods. Our proposed method is illustrated in the ~\ref{sec:Method}. Experiment setup and implementation details are in Section~\ref{sec:Experiment}. Results of the experiment and the discussion are in Section~\ref{sec:Results}. Finally, our conclusion about the result is in Section~\ref{sec:Conclusion}


\section{Related Work}
\label{sec:RelatedWork}

\subsection{Medication Recommendation}

In recent works, the categorization of medication recommendations involves two main approaches: instance-based and longitudinal recommendation methods, as highlighted by Shang et al. (2019) \cite{bhoi2020premier}. However, Jiang et al. (2016) \cite{hoens2013reliable} highlight the growing concern about health and medical diagnosis issues, emphasizing that medication errors, causing more than 1 crore deaths annually, are a significant problem. Novice doctors contribute to over 42\% of these errors due to limited experience and gaps in their knowledge. Data mining and recommender technologies offer a solution to leverage diagnosis history for accurate medication prescriptions and error reduction. On the other hand, the study of Jiang et al. (2016) \cite{hoens2013reliable} success heavily relies on accurate diagnosis history and data availability. The chosen Support Vector Machine (SVM) \cite{min2005recommender} model might have limitations in addressing certain complexities present in medical data.

\subsubsection{Instance-based Methods} This methodology exclusively relies on the current patient visits to generate medication recommendations. Zhang at el (2017) \cite{zhang2017leap} illustrate in their study outlined frames the challenge of medication recommendation using a multi-instance multi-label learning approach, introducing the LEAP algorithm. It incorporates a recurrent decoder to facilitate sequential decision-making, effectively modeling the correlation between drugs and diseases while considering DDI derived from EHR data. Once trained on patients' EHR data, this model predicts suitable medications based on a patient's existing diagnosis during their current visit.

Wang et al. (2017) \cite{wang2017safe} adopt a different approach by jointly embedding patient demographics, diagnosis, and historical medication records into a more compact dimensional space. This embedding is then utilized for making recommendations. They conceptualize drug recommendation as a link prediction task, leveraging patient diagnoses and potential adverse drug reactions.

Additionally, the research in \cite{wang2018personalized} discussed also leverages patient demographics, diagnoses, and medication details, integrating them through a trilinear method. However, these strategies overlook crucial patient information such as past diagnoses from earlier visits. As a consequence, there is a compromise in accuracy and a lack of personalization in the recommendations provided by these models.

\subsubsection{Longitudinal
Recommendation Methods}
In this healthcare context, recommender systems aim to harness the temporal dependencies inherent in a patient's historical medical data. In \cite{sun2016data}, a methodology is introduced that involves clustering patients' past medical records into cohorts based on treatment similarities, enabling the identification of patterns in treatment regimens.

Bajor and Lasko's work \cite{bajor2016predicting} suggests the use of Recurrent Neural Networks (RNNs) to predict medications based on a patient's clinical history and billing codes within EHR. However, their model is constrained to predicting whether a patient is using a specific drug and does not extend to recommending multiple medications, especially for complex medical conditions.

Shang et al. introduce a system known as GameNet \cite{shang2019gamenet}, which takes into account patients' longitudinal visit history and the complexities of drug interactions. Their approach incorporates dynamic memory and Graph Convolutional Networks (GCN) to personalize medication recommendations. However, an important limitation of GCN is its assumption that interactions between different drugs have uniform weights, which often doesn't hold true in practice. For instance, the severity of drug interactions can vary significantly. For instance, the adverse event of long-term or permanent paralysis resulting from the simultaneous use of an anti-inflammatory medication like Ibuprofen and an anticoagulant medication like Enoxaparin is considerably more severe compared to milder interactions, such as experiencing diarrhea after taking Ibuprofen alongside a constipation medication like Linaclotide.

\subsubsection{Attention-based Methods} Another noteworthy approach, as presented in \cite{choi2016retain}, with the capability to perform tasks such as disease and medication prediction. Building upon this foundation, Le et al. \cite{le2018dual} make further strides in the field by introducing DMNC, a model that combines memory-augmented neural networks with RNNs to effectively address long-range dependencies in medication recommendation. It's worth noting that these approaches, however, do not explicitly account for drug interactions in their methodologies.

\subsubsection{Multi-Task Learning Method} Inspired by the work of Yinying Zhang \cite{zhang2023knowledge}, proposing MedRec, an automated medicine recommendation system designed to address the challenge of sparse medical data. MedRec incorporates two distinct graphs for modeling, including knowledge-graph and attribute graph. These two graphs significantly enhance the connections between symptoms and medicines, effectively mitigating the issue of data sparsity. By leveraging these graph structures, MedRec learns the complex interrelationships between diseases, medicines, symptoms, and medical examinations, as well as the intrinsic associations within medicines themselves. Notably, our approach, ALGNet, also employs LGCN - parallel layers for efficiently processing medical relationship in numerous of layers. LGCN incorporates the use of low-weight graphs, resulting in reduced memory consumption compared to other methods. This optimization enhances the overall performance and scale opportunities of our recommendation system.

\begin{figure*}
\includegraphics[width=15cm]{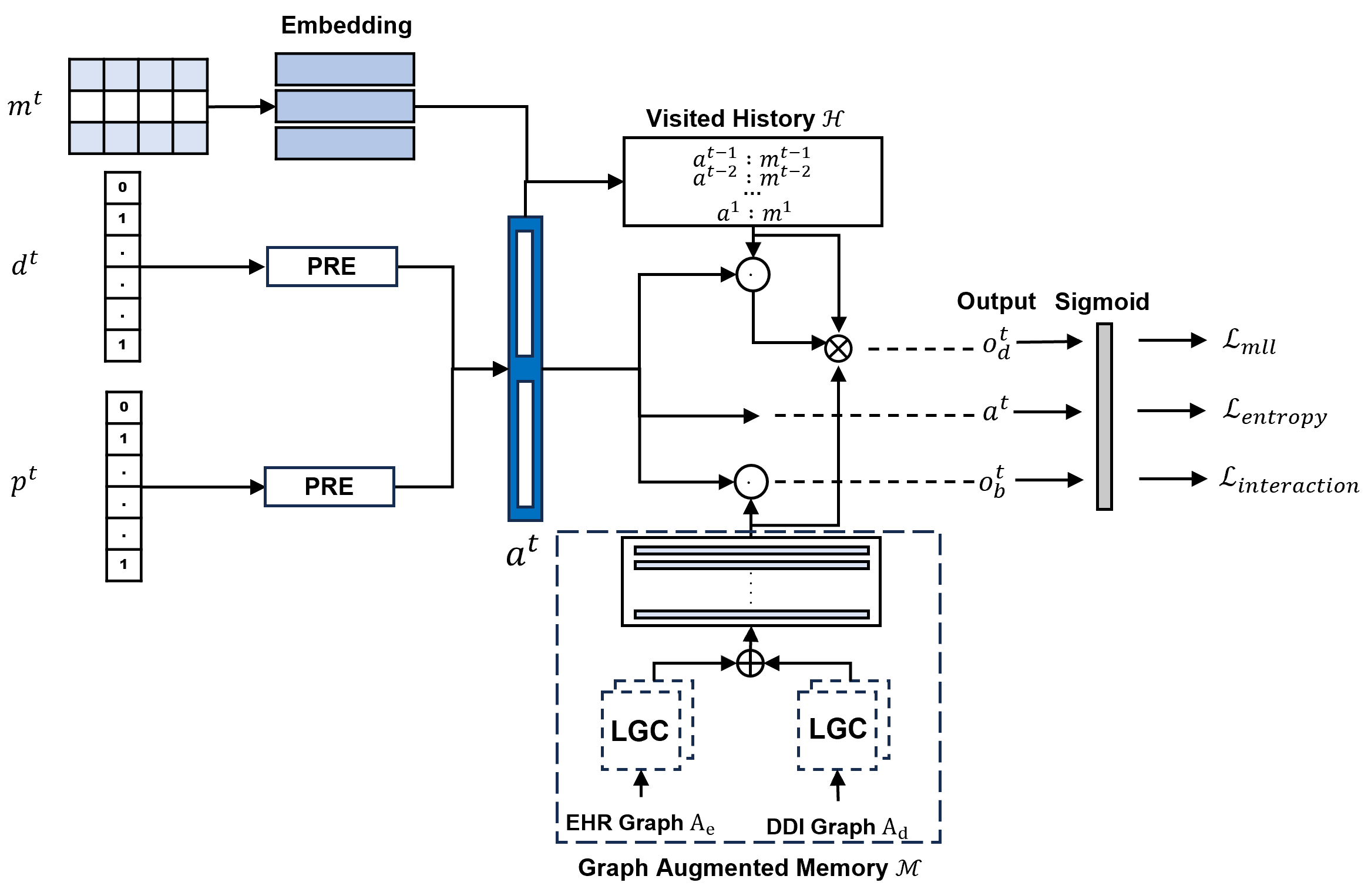}
\caption{The architecture of ALGNet. The PRE module is shown in Figure \ref{fig2} and the LGC layer is shown in Figure \ref{fig3}. In our works, we use three inputs from the patient (\ref{patient}) and two graph knowledge from EHR and DDI data (\ref{ehrddi}). The patient's inputs are walked through the PRE to be the embedding for the Equation \ref{eq8}. The other two graphs are used to make the memory graph by LGC layers and result in Equation \ref{eq2} and \ref{eq3}. The visited history is made by the medication embedding from the patients (Eq. \ref{eq9}). Then, the output is generated by Equation \ref{eq10} and \ref{eq11}, with a combination in Equation \ref{eq12}} \label{fig1}
\end{figure*}

\subsection{Light Graph Convolutional Networks}

 LGCN \cite{he2020lightgcn} is a novel technique for graph representation learning that aims to simplify the design and implementation of GCN \cite{zhang2019graph}. LGCN leverages the graph structure and node features to learn node embeddings without using any nonlinear activation functions, feature transformations, or weight matrices. LGCN can effectively capture the high-order connectivity and collaborative signals in user-item bipartite graphs, leading to more accurate and interpretable recommendations. LGCN has been applied in many recent recommendation systems with competitive results as \cite{huang2023dual}, \cite{ran2022pm}, and \cite{hansel2022optimized}.

\section{Proposed Method}
\label{sec:Method}
\subsection{Problem Statement}

In this section, we present our proposed method in detail. Figure \ref{fig1} shows an overview of our model architecture.

Our problem input can be defined by two main inputs of data, the Patient Records and the EHR \& DDI Graph. The goal of this problem is to suggest a set of drugs for a patient based on their current and past medical information and drug knowledge. The inputs are the diagnosis and procedure codes of the current visit, the previous visits of the patient, and two graphs that represent drug usage and drug interactions. The output is a multi-label vector that indicates which drugs are recommended.

To achieve this goal, our model consists of three main components: (1) a light graph convolutional network (LGCN) that encodes the patient record and the DDI graph into low-dimensional embeddings; (2) an augmentation memory network (AMN) that augments the patient representation with external knowledge from a memory module; (3) a decoder that generates the recommended medications based on the augmented patient representation.

\begin{table}
\caption{Example of a patient data record with two visits. The diagnosis, procedures, and medication are preprocessed from the original ICD code.}\label{tabdata}
\centering
\begin{tabular}{|l|l|l|}
\hline
& Visit 1 & Visit 2 \\
\hline
Diagnosis & 0, 1, 2, 3, 4, 5, 6, 7 & 8, 9, 10, 7 \\
\hline
Procedures & 0, 1, 2 & 3, 4, 1 \\
\hline
Medication & 0, 1, 2, 3, 4, 5, 6, & 0, 1, 2, 3, 5, 4, 6, 7, \\
& 7, 8, 9, 10, 11, 12, 13 & 8, 9, 10, 11, 13, 14, 15 \\
\hline
\end{tabular}
\end{table}

\subsubsection{The Patient Records} Patient records are sequences of multivariate observations that include diagnosis codes, procedure codes, and medication codes for each visit. Longitude EHR data includes multiple patient records in sequences \(EHR = \{x_1, x_2, ... x_i\}\). By this form, each patient includes a sequence of visits \(x_i=\{v_1, v_2, ..., v_t\}\). To infer it by formula, we denote the visit at time t of a patient is
$$  v_t = [d^t, p^t, m^t] $$  \label{patient}
where \(v_t\) 
is data of the visit at time \(t\) of a patient (\(t \in {1,2,...,n} \)) where \(n\) is the number of times that a patient visited. Consider that we only denote one patient in the sequences, the others corresponding to diagnosis codes \(d^t\), procedure codes \(p^t\), and medication codes \( m^t \) of a patient. Then, the medical code set is denoted as \(\mathcal{C}_*\) corresponding to \( \mathcal{C}_d, \mathcal{C}_p, \mathcal{C}_m  \). As the point of that, \(|\mathcal{C}_*|\) is the number of each medical code type. We can view an example of a patient data as Table \ref{tabdata}.

\subsubsection{EHR \& DDI Graph} \label{ehrddi}

EHR graph is constructed by an adjacency matrix \( A_e \in \mathbb{R}^{|\mathcal{C}_m| \times |\mathcal{C}_m|} \) to indicate the number of drug co-occurrences, which is the relation of medication in the EHR records. In other words, it represents the medical combinations that have appeared in the patient records at the previous visits. On the other side, we constructed the DDI adjacency matrix \( A_d \in \mathbb{R}^{|\mathcal{C}_m| \times |\mathcal{C}_m|} \) from the drugs-drugs interaction graph database \cite{tatonetti2012data}. In this graph, each connection represents the interaction of a pair of drugs.

\subsection{Light Graph Convolution}

We illustrate the two layers of Light Graph Convolution (LGC) as in Figure \ref{fig2}.

\begin{figure}
\includegraphics[width=3cm]{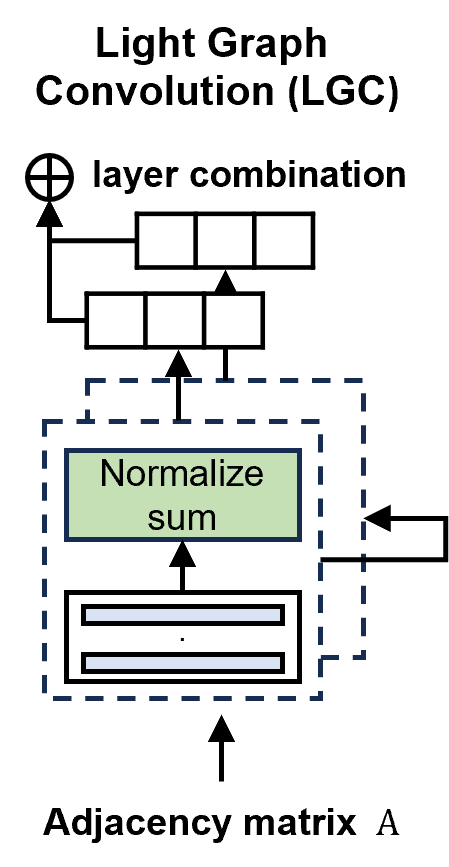}
\caption{Light Graph Convolutional Layer (LGC)} \label{fig3}
\end{figure}

Firstly, to build an augmented memory network, we must implement the memory graph data, which consists of EHR and DDI knowledge. Each of the graph inputs is embedded by the graph convolutional layer from \cite{he2020lightgcn}. 

\begin{equation}\label{eq1}
    E^{k+1} = (D^{-\frac{1}{2}} A D^{-\frac{1}{2}}) E^k  
\end{equation}

where \( \tilde{A} = D^{-\frac{1}{2}} A D^{-\frac{1}{2}} \) is the symmetrically normalized matrix of \(A\), \( E^k \) is the graph at the layer \(k^{th}\). At layer 0, \(A\in\{A_e, A_d\}\). After 2 layers, We combined all layers to get the final embedding of each graph, then summed all weights of two embedding from EHR and DDI graph embedding to get the memory graph \(M\).

\begin{equation}\label{eq2}
    E= \alpha E_1+\alpha E_2 
\end{equation}

\begin{equation}\label{eq3}
    \mathcal{M}= E_{A_e} + \beta E_{A_d}
\end{equation}

where \( \alpha \) controls the importance of each layer and \( \beta \) is a weighting variable to interact with the drugs knowledge graph. By using this light graph, we can skip the non-linear transformation and remove the dropout, the computation cost will be reduced but still keep the features in the graph.

\subsection{Input Embedding}

\begin{figure}
\includegraphics[width=6cm]{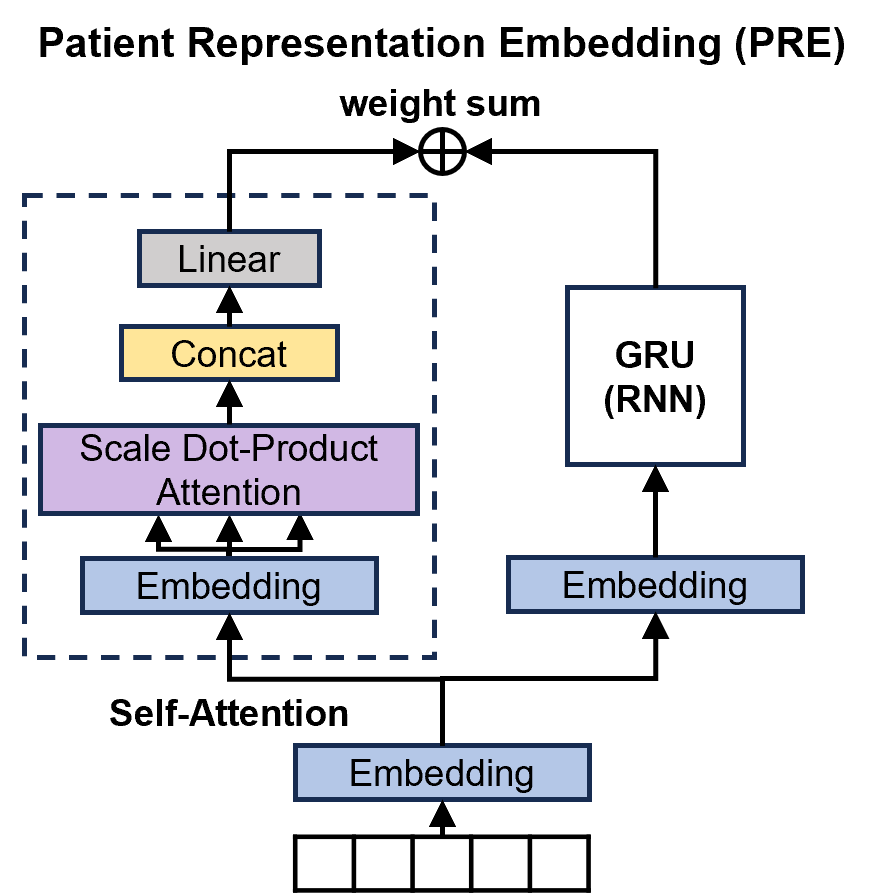}
\caption{Details of Patient Representation Embedding (PRE)} \label{fig2}
\end{figure}

The input of this network is a patient record, which includes \( d^t, p^t, m^t \). We would take the diagnosis \( d^t \) and procedure \( p^t \) through the dual patient representation embedding (PRE) layer, about past medical records, they are used to construct the visited history \( \mathcal{H} \). Each layer consists of two main extractors: RNN and Self-Attention. Firstly, all three inputs are embedded into a linear embedding layer to a 64-dimensional space, which is to learn the distributed representations, by using the embedding matrix \( d^t \rightarrow e_d^t \in \mathbb{R}^{|\mathcal{C}_d| \times 64}, p^t \rightarrow e_p^t \in \mathbb{R}^{|\mathcal{C}_e| \times 64}, m^t \rightarrow e_m^t \in \mathbb{R}^{|\mathcal{C}_m| \times 64} \). The details of this embedding are illustrated in Fig. \ref{fig2}.

\subsubsection{Multi-Head Self-Attention}

From Fig.~\ref{fig1}, we consider taking more information about the long-range medical records and learning more features from the patient's current status, so a new self-attention module was added to retrieve these features and then combined with the RNN module. The main architecture of the self-attention module here from \cite{vaswani2023attention} where our inputs \( e_* \) (\(* \in \{d,p\}\)), which consist of all patient previous visits' diagnosis and procedures, are mapping to queries \( Q_* = e_*W^Q\), keys  \( K_*=e_*W^K \), and values  \( V_*=e_*W^V \). Then, all of that is passed through the multi-head attention before concatenating and linear transforming. For each parameter matrix, we also use 64 as the number of dimensions to project these values to the 64-dimensional space and capture full features from the previous embedding.

\begin{equation}
    Attention(Q,K,V) = softmax(\frac{QK^T}{\sqrt{di_k}})V
\end{equation}

\begin{equation}
    Multihead(Q,K,V) = concat(h_1, h_2, ..., h_h)W^O 
\end{equation}

   $$ where \hspace{0.2cm} h_i = Attention(Q_i,K_i,V_i) $$

Where \(di_k=64\) is the dimension of keys and \( W^Q,W^K,W^V \in \mathbb{R}^{di_{model} \times 64} \); \( W^O \in \mathbb{R}^{h \times di_{model} \times 64} \). For each of these, we use \(di_{model}=64*h\). In our work, we set the number of heads \(h =8 \). Finally, the results after concatenation are walked through a linear embedding layer to match the dimension of the results from the RNN module, we denote this stage result is \( s_{e_*} \)

\subsubsection{RNN Module (GRU)}

To take advantage of personalized medical recommendations from the patient EHR data, we inherit the RNN module for diagnosis and procedure modalities separately. For each patient, we retrieve all previous visits and utilize the RNN to encode the patient diagnosis and procedure embedding.

\begin{equation}
    l_{e_*} = RNN(e_*^1, e_*^2, ... e_*^t)
\end{equation}

Where \(l_{e_*}\) is a hidden layer resulting from the RNN module and used for further calculating.

\subsubsection{Weight sum and Concatenate}

Both results from the Multi-Head Self-Attention and RNN module are weighted sum and produce a final embedding of each type (diagnosis and procedure). Then we concatenate both the diagnosis and procedure stages to represent the patient's current stage by a transform function, which is a set of hidden layers and a fully connected neural network layer. The final representation of the PRE module is:

\begin{equation}
    PRE(e_*) = l_{e_*} + \gamma s_{e_*}
\end{equation}

where \(\gamma\) is used for driving the ratio of the multi-head attention module. To generate the embedding for the next step, we denote:

\begin{equation} \label{eq8}
    a^t = f(PRE(e_d^t), PRE(e_p^t))
\end{equation}

where \(f(\cdot)\) is the transform function.

\subsection{Augmentation Memory Network}

The augmentation memory network architecture is shown in (Fig.~\ref{fig1}). After the two light graphs are computed, all these two are used to generate the memory graph \(\mathcal{M}\) from the Eq.~\ref{eq3}

For the visited history memory of a patient \(\mathcal{H}\), we combine each previous medication embedding \(m^t\) with its correlative patient representation embedding \(a^t\) as a key-value pair.

\begin{equation}\label{eq9}
    \mathcal{H} = \{ a^{i} : m^{i}; 0<i<t, i \in \mathbb{N} \}
\end{equation}

By this formula, \(\mathcal{H}\) is empty when \(t=1\). To continue with the next step, we can represent the keys and values of \(\mathcal{H}\) as \( \mathcal{H}_k = [a^1, a^2, ..., a^{t-1}] \in \mathbb{R}^{|t-1| \times 64} \) and \( \mathcal{H}_v = [m^1, m^2, ..., m^{t-1}] \in \mathbb{R}^{|t-1| \times |\mathcal{C}_m|} \). Note that all these denotes are used for the visit at time \(t^{th}\).

\subsection{Output representation}

We use two outputs generated from both memories \(\mathcal{H}\) and \(\mathcal{M}\) with the patient representation embedding \(a^t\).

\begin{equation}\label{eq10}
    o^t_b = \mathcal{M}^\mathbf{T} Softmax( \mathcal{M}a^t )
\end{equation}

\begin{equation}\label{eq11}
    o^t_d = \mathcal{M}^\mathbf{T} \mathcal{H}_v^\mathbf{T} Softmax( \mathcal{H}_k a^t )
\end{equation}

where \( o^t_b \in \mathbb{R}^{|\mathcal{C}_m| \times 64} \), \( o^t_d \in \mathbb{R}^{|\mathcal{C}_m| \times |t-1|} \)

To provide the output recommendation, we combine the outputs with the patient representation. The prediction at the \(t^{th}\) visit of a patient is as follows:

\begin{equation}\label{eq12}
    \hat{y}^t=\sigma (a^t, o^t_b, o^t_d)
\end{equation}

where \(\sigma\) is the sigmoid function.

\subsection{Loss function}

Currently, we use two common losses for this recommendation task multi-label loss (MLL) and binary cross-entropy loss to compute and optimize the prediction results. \( \hat{y}^t[i], \hat{y}^t_i \) is denoted as the value at \( i^{th} \) index and visit number t.

\begin{equation}
    \mathcal{L}_{mll} = \sum^{T}_{t}{\sum^{|\mathcal{C}_m|}_{i}{\sum^{|\hat{Y}^t|}_{j}{ \frac{ max(0,1-(\hat{y}^t[\hat{Y}^t_j] - \hat{y}^t[i])) }{L} }}}
\end{equation}

where $\hat{Y}^t=\{ \hat{y}^t_j | \hat{y}^t_j > 0.5, 1 \leq j \leq | \mathcal{C}_m | \} $
is the prediction set with threshold = 0.5.

\begin{equation}
    \mathcal{L}_{entropy} = - \sum^{T}_{t}{\sum_{i}{ y^t_i \mathrm{log}\sigma(\hat{y}^t_i) + (1 - y^t_i) \mathrm{log}(1 - \sigma(\hat{y}^t_i)) } }
\end{equation}

We do not focus on the DDI rate but still keep it low, not increasing gradually by an interaction loss from \cite{shang2019gamenet}.

\begin{equation}
    \mathcal{L}_{interaction} = \sum^{T}_{t} \sum_{i,j} (A_d \odot \hat{y}^{\mathrm{T} }_t  \hat{y}_t)[i,j]
\end{equation}

To control the DDI Rate and not let it increase too much, we use a combined loss function from \cite{shang2019gamenet} for the total loss with the interaction loss that helps the loss active and also keeps the DDI loss not affecting too much to the total loss. For the multi-label loss and binary cross-entropy loss, we use a set of weighted variables \( \{\theta_0, \theta_1 \} \) to drive the importance of each loss, where the total of all values in the set is 1. This variable would be tuned during the training process.

\[ \mathcal{L} = \theta_0 \mathcal{L}_{entropy} + \theta_1 \mathcal{L}_{mll} \]

The training algorithm is detailed as follows.
\begin{algorithm}
	\caption{Training ALGNet} 
	\begin{algorithmic}[1]
        \State \textbf{Require:} Training set R, training epochs N,$\alpha$ ratio in Eq. 2, mixture weight $\beta$ in Eq. \ref{eq3}, expected DDI Rate $s$, $\gamma$ ratio in Eq. 7;
        \State Calculate adjacency matrix $A_{*}$;
        \State Obtain Memory Graph $\mathcal{M}$ using Eq. \ref{eq3};
		\For {$iteration=1$} to $N*|R|$ do
            \State Sample a patient P =[$x_1,x_2,···,x_{T_i}$] and time visit $v_t=[d^t, p^t, m^t]$ from R
            \State Reset Visited History Memory $\mathcal{H}$;
			\For {$T=1$} to $T_i$ do
                \State Obtain medical embedding $e^t_d , e^t_p, e^t_m$ as mentioned in section 3.3;
                \State Apply Multi-head Attention to $e^t_d , e^t_p$ in Eq. 4 and 5
				\State Obtain RNN Module (GRU) $l_e$ in Eq. 6;
                \State Compute the $PRE(e_{*})$ of $e^t_d , e^t_p$ in Eq. 7;
				\State Compute the concatenate of $a^t=f(PRE(e^t_d),PRE(e^t_p))$
                \State Read from $\mathcal{M}$ in Eq. \ref{eq3} and $\mathcal{H} =[m^1, m^2,\ldots,m^{t-1}]$ in Eq. 9;
                \State Generate output of $o^t_b$ and $o^t_d$ using Eq. 10 and 11
                \State Calculate medication prediction $\hat{y}^t$ using Eq. 12
                \State Generate $\mathcal{H}^{t+1}$ by inserting $(a^t, e^t_m)$ into $\mathcal{H}^t$.
			\EndFor
			\State Evaluate and obtain DDI Rate $s'$ of current patient using loss Eq. 15
			\State Update all $\mathbf{W}$ by optimizing loss in Eq. 13 and 14;
		\EndFor
	\end{algorithmic} 
\end{algorithm}

\begin{table}
\caption{Statistics of the Data}\label{tab1}
\centering
\begin{tabular}{|l|l|}
\hline
Field & Records \\
\hline
No. Patients & 6.350 \\
No. Clinical Events & 15.016 \\ 
No. Diagnosis & 0.0949 \\
No. Procedure & 1,426 \\
No. Medication & 145 \\
\hline
No. Avg of Visit & 2.36 \\
No. Avg of Diagnosis & 10.51 \\
No. Avg of Procedure & 3.84 \\
No. Avg of Medication & 8.80 \\
\hline
\end{tabular}
\end{table}

\begin{table*}[ht]
\caption{Results on MIMIC-III dataset}\label{tab2}
\centering
\begin{tabular}{|c|c|c|c|c|c|}
\hline
Model & DDI Rate &  Jaccard & F1-Score & PR-AUC & Avg. \# of Drugs\\ 
\hline
Leap & 0.0731 \(\pm\) 0.0008 & 0.4521 \(\pm\) 0.0024 & 0.6138 \(\pm\) 0.0026 & 0.6549 \(\pm\) 0.0033 &  20.4051 ± 0.2832 \\
RETAIN & 0.0835 ± 0.0020 & 0.4887 ± 0.0028 & 0.6481 ± 0.0027 & 0.7556 ± 0.0033 & 18.7138 ± 0.0666 \\ 
DMNC & 0.0842 ± 0.0011 & 0.4864 ± 0.0025 & 0.6529 ± 0.0030 & 0.7580 ± 0.0039 & 20.0000 ± 0.0000 \\
GAMENet & 0.0864 ± 0.0006 & 0.5067 ± 0.0025 & 0.6626 ± 0.0025 & 0.7631 ± 0.0030 &  27.2145 ± 0.1141 \\
\hline
SafeDrug & \textbf{0.0598 ± 0.0005} & 0.5113 ± 0.0030 & 0.6680 ± 0.0027 & 0.7663 ± 0.0025 & 19.9178 ± 0.1604 \\
MICRON &  0.0678 ± 0.0007 & 0.5100 ± 0.0033 & 0.6654 ± 0.0031 & 0.7687 ± 0.0026 & 17.9267 ± 0.2172 \\
\hline
\textbf{Our works (ALGNet)} & 0.0791 ± 0.0012 & \textbf{0.5176 ± 0.0010} & \textbf{0.6729 ± 0.0007} & \textbf{0.7714 ± 0.0043} & 25.8400 ± 0.0017 \\
\hline
\end{tabular}
\end{table*}

\section{Experiment}
\label{sec:Experiment}
\subsection{Datasets}
The performance evaluation of the ALGNet model involves comparison and reference with other baseline models in terms of recommendation accuracy and the effective avoidance of DDI. Our evaluation utilizes the MIMIC-III \cite{johnson2016mimic} database, as documented in Johnson et al.'s work (2020) \cite{johnson2020mimic}. This database contains critical care data for more than 40,000 patients admitted to intensive care units at the Beth Israel Deaconess Medical Center (BIDMC). In our study, we focus on medications prescribed within the crucial initial 24-hour window, a period crucial for swift and precise patient treatment. These findings are succinctly presented in Table~\ref{tab1}.

\subsection{Implementation details}

\subsubsection{Evaluation Metrics}
As an evaluation in other baselines of SafeDrug \cite{yang2022safedrug}, we employ a set of four efficacy metrics, namely the DDI rate, Jaccard similarity, F1 score, and PRAUC (Precision Recall Area Under Curve), to assess the effectiveness of the recommendations. Additionally, in this work, our target is to improve the accuracy of 3 trivial prediction metrics, which are Jaccard, PR-AUC, and F1 Score. In which Jaccard is defined as the size of the intersection divided by the size of the union of ground
truth medications $Y_t^{(k)}$ and predicted medications $\hat{Y}_t^{(k)}$.

$$
Jaccard = \frac{1}{\sum_{k}^{N} \sum_{T_k}^{t} 1} \sum_{N}^{k} \sum_{T_k}^{t} \frac{|Y_t^{(k)} \bigcap \hat{Y}_t^{(k)}|}{|Y_t^{(k)} \bigcup \hat{Y}_t^{(k)}|}
$$

where $N$ is the number of patients in test set and $T_k$ is the
number of visits of the $k^{th}$ patient. Average Precision (Avg- P) and Average Recall (Avg-R), and F1 are defined as:

$$
\text{Avg-P}_t^{(k)} = \frac{|Y_t^{(k)} \bigcap \hat{Y}_t^{(k)}|}{Y_t^{(k)}}
$$

$$
\text{Avg-R}_t^{(k)} = \frac{|Y_t^{(k)} \bigcap \hat{Y}_t^{(k)}|}{\hat{Y}_t^{(k)}}
$$

$$
F1 = \frac{1}{\sum_{k}^{N} \sum_{T_k}^{t} 1} \sum_{N}^{k} \sum_{T_k}^{t} \frac{2 \times \text{Avg-P}_t^{(k)} \times \text{Avg-R}_t^{(k)}}{\text{Avg-P}_t^{(k)} + \text{Avg-R}_t^{(k)}}
$$

where $t$ means the $t^{th}$ visit and $k$ means the $k^{th}$ patient in test dataset.

\subsubsection{Training Setup}
Our experimental setup aligns with that of \cite{shang2019gamenet}. The dataset is divided into training, validation, and test sets in a $\frac{2}{3}: \frac{1}{6}: \frac{1}{6}$ ratio respectively, randomly partitioned. Models are implemented in PyTorch and parameters are trained on Adam optimizer \cite{kinga2015method} with a learning rate $2*10^{-4}$ for around 40-80 epochs. We use a Ubuntu environment with configuration: environment of version 22.04 and one Nvidia RTX 3060 GPU to implement the experiment.

\section{Results}
\label{sec:Results}





\subsection{Comparative Results}

To evaluate the effectiveness of our method, we do the benchmark of our method with these relevant baselines: LEAP: \cite{zhang2017leap}, RETAIN: \cite{choi2016retain}, DMNC: \cite{le2018dual}, GAMENet (2019) \cite{shang2019gamenet}, SafeDrug (2021) \cite{yang2022safedrug},  MICRON (2021) \cite{yang2021change}.

As the report of baseline, Table~\ref{tab2} provides a comprehensive performance comparison for accuracy and safety aspects within the MIMIC-III dataset. ALGNet consistently outperforms all baseline models in terms of Jaccard, PR-AUC, and F1 scores. 

Among the baseline models, RETAIN and DMNC exhibit roughly a 3\% lower performance compared to ALGNet in terms of Jaccard and F1 scores. Notably, Leap demonstrates a significantly larger performance gap in terms of Jaccard. The Nearest method underscores the importance of patient visit history in medication combination recommendations.

Conversely, in the case of more recent methods like SafeDrug (2021) and MICRON (2021), they indeed attain higher scores in terms of Jaccard, PRAUC, and F1 when compared to other models. These methods recommend a substantial number of medication combinations, which may contribute to the elevated DDI Rate observed. However, it's worth noting that ALGNet maintains its superiority in terms of DDI rate, showcasing a range of 5-10\% improvement, and exhibits a 0.5-1\% edge in other relevant metrics.

In the field of DDI rates, our model did not target lower this rate. However, we still maintained this rate did not increase rapidly and kept it lower than some other baselines such as RETAIN, DMNC, and GAMENET. At this rate, our model can be accurate and not create biased results.

In conclusion, our work, ALGNet, exhibits clear advantages over existing recommendation systems in healthcare. Its lower DDI Rate, higher Jaccard Index, balanced F1-Score, superior PR-AUC, and reasonable average number of drugs recommended collectively make it a compelling choice for healthcare professionals seeking to optimize drug recommendations. ALGNet's ability to enhance patient safety, accuracy, and efficiency in decision-making sets it apart as a promising solution for personalized healthcare recommendations.

\subsection{Qualitative Result}

The evaluation of our method on the MIMIC-III dataset is demonstrated in Table~\ref{tab2}, with the highest values shown in \textbf{bold}. As can be observed, our method achieves competitive results compared to previous methods. In terms of the Jaccard metric, ALG-Net exhibits a performance improvement of $+0.0063$. Regarding the F1-score, our method enhances the result by more than $+0.0049$ compared to the previous method. In the case of the PR-AUC metric, our method surpasses the second-place competitor by $+0.0028$.

These results illustrate that our proposed model can effectively capture a better data representation for the recommendation task. Moreover, the Light-GCN method efficiently captures the graph signal, resulting in significant performance improvements when training on a small-scale dataset. Consequently, our model achieves competitive results compared to previous approaches.

\subsection{Ablation Study}

In this section, we conducted an experiment where we removed the Light Graph Convolution component from ALGNet, denoted as ALGNet w/o $E^{k+1}$, to assess its impact on model performance.

The results presented in Table~\ref{tab3} and Table~\ref{tab4} clearly show that the removal of LGCN has a discernible effect on ALGNet's performance. This indicates the crucial role played by this graph in representing medications effectively.

\begin{table}[H]
\caption{Results of each embedding type without LGC}\label{tab3}
\centering
\begin{tabular}{|l|l|l|l|l|}
\hline
Model & DDI Rate &  Jaccard & PR-AUC & F1 Score\\
\hline
LSTM-GCN &  0.0917 & 0.5018 &  0.7521 & 0.6473 \\
RNN-GCN (GAMENet) & \textbf{0.0864}  & 0.5067 & 0.7631 & 0.6626 \\
A-GCN & 0.0869 & 0.5076 & 0.7632 & 0.6628 \\
A-LSTM-GCN & 0.0886 & 0.5068 &  0.7639 & 0.6632 \\
A-RNN-GCN & 0.0892 & \textbf{0.5091} & \textbf{0.7650} & \textbf{0.6645} \\
\hline
\end{tabular}
\end{table}

Furthermore, the inclusion of an optimized graph theory structure also significantly contributes to the effectiveness of medication representation. When we eliminate the optimized graph mechanism from the ALGNet model, we observe a notable decline in performance. This underscores the critical role of the LtGCN mechanism within the model.

Additionally, the outcomes achieved by employing RNN or LSTM models highlight the superiority of the Transformer encoder in capturing essential information from patients' visit records and medication history from a holistic perspective. This underscores the significance of graph-based collaborative decision-making in enhancing the model's overall effectiveness.

In summary, these findings underscore the importance of incorporating the Light Graph Convolution Network into ALGNet to achieve superior performance in medication recommendation tasks.

\begin{table}[H]
\caption{Results of each embedding type with LGC}\label{tab4}
\centering
\begin{tabular}{|l|l|l|l|l|}
\hline
Model & DDI Rate &  Jaccard & PR-AUC & F1 Score\\
\hline

LSTM-LGNet & 0.0831 & 0.5021 &  0.7531 & 0.6487 \\
RNN-LGNet & 0.0810 & 0.5071 & 0.7641 & 0.6636 \\
A-LGNet\(^{-RNN}\) & \textbf{0.0779} & 0.5112 & 0.7653 & 0.6668 \\
A-LSTM-LGNet & 0.0860  & 0.5122 &  0.7681 & 0.6671 \\
\textbf{ALGNet} & 0.0791 & \textbf{0.5176} & \textbf{0.7714} & \textbf{0.6729} \\
\hline
\end{tabular}
\end{table}

\section{Conclusion}
\label{sec:Conclusion}

In this paper, we have proposed ALGNet, a novel model that leverages light graph convolutional networks and augmentation memory networks to enhance medication recommendation. ALGNet can efficiently encode the patient records and the DDI graph into low-dimensional embeddings, while augmenting the patient representation with external knowledge from a memory module. We have evaluated our model on the MIMIC-III dataset and shown that it outperforms several baselines in terms of recommendation accuracy and DDI avoidance. We have also conducted an ablation study to analyze the effects of different components of our model. Our results demonstrate that ALGNet can achieve superior performance with less computation and more interpretability.

Our work has several implications for the field of medical recommendation systems. First, it shows that graph neural networks can effectively capture the complex and dynamic relationships among patient medical records, drug efficacy and safety, and drug-drug interactions. Second, it shows that memory-augmented neural networks can enrich the patient representation with external knowledge from various sources, such as drug labels, medical ontologies, or literature. Third, it shows that a simple feed-forward network can generate accurate and safe medication recommendations based on the augmented patient representation.

Our work also opens up some directions for future research. First, we plan to extend our model to handle multi-modal data, such as images, texts, or sounds, that may provide additional information for medication recommendation. Second, we plan to incorporate more domain knowledge and constraints into our model, such as dosage, frequency, duration, or contraindications. Third, we plan to explore more advanced architectures and techniques for our model, such as transformer networks, graph attention networks, or reinforcement learning.

We hope that our work can inspire more research on medication recommendation and contribute to improving patient care and reducing adverse events.

\begin{acks}
\label{sec:Acknowledgement}
This research is supported by research funding from Faculty of Information Technology, University of Science, Vietnam National University - Ho Chi Minh City.
\end{acks}

\bibliographystyle{ACM-Reference-Format}
\bibliography{sample-base}


\begin{thebibliography}{29}


\ifx \showCODEN    \undefined \def \showCODEN     #1{\unskip}     \fi
\ifx \showDOI      \undefined \def \showDOI       #1{#1}\fi
\ifx \showISBNx    \undefined \def \showISBNx     #1{\unskip}     \fi
\ifx \showISBNxiii \undefined \def \showISBNxiii  #1{\unskip}     \fi
\ifx \showISSN     \undefined \def \showISSN      #1{\unskip}     \fi
\ifx \showLCCN     \undefined \def \showLCCN      #1{\unskip}     \fi
\ifx \shownote     \undefined \def \shownote      #1{#1}          \fi
\ifx \showarticletitle \undefined \def \showarticletitle #1{#1}   \fi
\ifx \showURL      \undefined \def \showURL       {\relax}        \fi
\providecommand\bibfield[2]{#2}
\providecommand\bibinfo[2]{#2}
\providecommand\natexlab[1]{#1}
\providecommand\showeprint[2][]{arXiv:#2}

\bibitem[Bajor and Lasko(2016)]%
        {bajor2016predicting}
\bibfield{author}{\bibinfo{person}{Jacek~M Bajor} {and}
  \bibinfo{person}{Thomas~A Lasko}.} \bibinfo{year}{2016}\natexlab{}.
\newblock \showarticletitle{Predicting medications from diagnostic codes with
  recurrent neural networks}. In \bibinfo{booktitle}{\emph{International
  conference on learning representations}}.
\newblock


\bibitem[Bhoi et~al\mbox{.}(2020)]%
        {bhoi2020premier}
\bibfield{author}{\bibinfo{person}{Suman Bhoi}, \bibinfo{person}{Lee~Mong Li},
  {and} \bibinfo{person}{Wynne Hsu}.} \bibinfo{year}{2020}\natexlab{}.
\newblock \showarticletitle{Premier: Personalized recommendation for medical
  prescriptions from electronic records}.
\newblock \bibinfo{journal}{\emph{arXiv preprint arXiv:2008.13569}}
  (\bibinfo{year}{2020}).
\newblock


\bibitem[Chen et~al\mbox{.}(2013)]%
        {chen2013human}
\bibfield{author}{\bibinfo{person}{Li Chen}, \bibinfo{person}{Marco De~Gemmis},
  \bibinfo{person}{Alexander Felfernig}, \bibinfo{person}{Pasquale Lops},
  \bibinfo{person}{Francesco Ricci}, {and} \bibinfo{person}{Giovanni
  Semeraro}.} \bibinfo{year}{2013}\natexlab{}.
\newblock \showarticletitle{Human decision making and recommender systems}.
\newblock \bibinfo{journal}{\emph{ACM Transactions on Interactive Intelligent
  Systems (TiiS)}} \bibinfo{volume}{3}, \bibinfo{number}{3}
  (\bibinfo{year}{2013}), \bibinfo{pages}{1--7}.
\newblock


\bibitem[Choi et~al\mbox{.}(2016)]%
        {choi2016retain}
\bibfield{author}{\bibinfo{person}{Edward Choi}, \bibinfo{person}{Mohammad~Taha
  Bahadori}, \bibinfo{person}{Jimeng Sun}, \bibinfo{person}{Joshua Kulas},
  \bibinfo{person}{Andy Schuetz}, {and} \bibinfo{person}{Walter Stewart}.}
  \bibinfo{year}{2016}\natexlab{}.
\newblock \showarticletitle{Retain: An interpretable predictive model for
  healthcare using reverse time attention mechanism}.
\newblock \bibinfo{journal}{\emph{Advances in neural information processing
  systems}}  \bibinfo{volume}{29} (\bibinfo{year}{2016}).
\newblock


\bibitem[Das et~al\mbox{.}(2017)]%
        {das2017survey}
\bibfield{author}{\bibinfo{person}{Debashis Das}, \bibinfo{person}{Laxman
  Sahoo}, {and} \bibinfo{person}{Sujoy Datta}.}
  \bibinfo{year}{2017}\natexlab{}.
\newblock \showarticletitle{A survey on recommendation system}.
\newblock \bibinfo{journal}{\emph{International Journal of Computer
  Applications}} \bibinfo{volume}{160}, \bibinfo{number}{7}
  (\bibinfo{year}{2017}).
\newblock


\bibitem[Fan et~al\mbox{.}(2019)]%
        {fan2019graph}
\bibfield{author}{\bibinfo{person}{Wenqi Fan}, \bibinfo{person}{Yao Ma},
  \bibinfo{person}{Qing Li}, \bibinfo{person}{Yuan He}, \bibinfo{person}{Eric
  Zhao}, \bibinfo{person}{Jiliang Tang}, {and} \bibinfo{person}{Dawei Yin}.}
  \bibinfo{year}{2019}\natexlab{}.
\newblock \showarticletitle{Graph neural networks for social recommendation}.
  In \bibinfo{booktitle}{\emph{The world wide web conference}}.
  \bibinfo{pages}{417--426}.
\newblock


\bibitem[Hansel et~al\mbox{.}(2022)]%
        {hansel2022optimized}
\bibfield{author}{\bibinfo{person}{Aaron~Christian Hansel},
  \bibinfo{person}{Lunardi Pradana}, \bibinfo{person}{Abba Suganda},
  \bibinfo{person}{Ariadi Nugroho}, {et~al\mbox{.}}}
  \bibinfo{year}{2022}\natexlab{}.
\newblock \showarticletitle{Optimized LightGCN for Music Recommendation
  Satisfaction}. In \bibinfo{booktitle}{\emph{2022 6th International Conference
  on Information Technology, Information Systems and Electrical Engineering
  (ICITISEE)}}. IEEE, \bibinfo{pages}{449--454}.
\newblock


\bibitem[He et~al\mbox{.}(2020)]%
        {he2020lightgcn}
\bibfield{author}{\bibinfo{person}{Xiangnan He}, \bibinfo{person}{Kuan Deng},
  \bibinfo{person}{Xiang Wang}, \bibinfo{person}{Yan Li},
  \bibinfo{person}{Yongdong Zhang}, {and} \bibinfo{person}{Meng Wang}.}
  \bibinfo{year}{2020}\natexlab{}.
\newblock \showarticletitle{Lightgcn: Simplifying and powering graph
  convolution network for recommendation}. In
  \bibinfo{booktitle}{\emph{Proceedings of the 43rd International ACM SIGIR
  conference on research and development in Information Retrieval}}.
  \bibinfo{pages}{639--648}.
\newblock


\bibitem[Hoens et~al\mbox{.}(2013)]%
        {hoens2013reliable}
\bibfield{author}{\bibinfo{person}{T~Ryan Hoens}, \bibinfo{person}{Marina
  Blanton}, \bibinfo{person}{Aaron Steele}, {and} \bibinfo{person}{Nitesh~V
  Chawla}.} \bibinfo{year}{2013}\natexlab{}.
\newblock \showarticletitle{Reliable medical recommendation systems with
  patient privacy}.
\newblock \bibinfo{journal}{\emph{ACM Transactions on Intelligent Systems and
  Technology (TIST)}} \bibinfo{volume}{4}, \bibinfo{number}{4}
  (\bibinfo{year}{2013}), \bibinfo{pages}{1--31}.
\newblock


\bibitem[Huang et~al\mbox{.}(2023)]%
        {huang2023dual}
\bibfield{author}{\bibinfo{person}{Wenqing Huang}, \bibinfo{person}{Fei Hao},
  \bibinfo{person}{Jiaxing Shang}, \bibinfo{person}{Wangyang Yu},
  \bibinfo{person}{Shengke Zeng}, \bibinfo{person}{Carmen Bisogni}, {and}
  \bibinfo{person}{Vincenzo Loia}.} \bibinfo{year}{2023}\natexlab{}.
\newblock \showarticletitle{Dual-LightGCN: Dual light graph convolutional
  network for discriminative recommendation}.
\newblock \bibinfo{journal}{\emph{Computer Communications}}
  \bibinfo{volume}{204} (\bibinfo{year}{2023}), \bibinfo{pages}{89--100}.
\newblock


\bibitem[Johnson et~al\mbox{.}(2020)]%
        {johnson2020mimic}
\bibfield{author}{\bibinfo{person}{Alistair Johnson}, \bibinfo{person}{Lucas
  Bulgarelli}, \bibinfo{person}{Tom Pollard}, \bibinfo{person}{Steven Horng},
  \bibinfo{person}{Leo~Anthony Celi}, {and} \bibinfo{person}{Roger Mark}.}
  \bibinfo{year}{2020}\natexlab{}.
\newblock \showarticletitle{Mimic-iv}.
\newblock \bibinfo{journal}{\emph{PhysioNet. Available online at:
  https://physionet. org/content/mimiciv/1.0/(accessed August 23, 2021)}}
  (\bibinfo{year}{2020}).
\newblock


\bibitem[Johnson et~al\mbox{.}(2016)]%
        {johnson2016mimic}
\bibfield{author}{\bibinfo{person}{Alistair~EW Johnson}, \bibinfo{person}{Tom~J
  Pollard}, \bibinfo{person}{Lu Shen}, \bibinfo{person}{Li-wei~H Lehman},
  \bibinfo{person}{Mengling Feng}, \bibinfo{person}{Mohammad Ghassemi},
  \bibinfo{person}{Benjamin Moody}, \bibinfo{person}{Peter Szolovits},
  \bibinfo{person}{Leo Anthony~Celi}, {and} \bibinfo{person}{Roger~G Mark}.}
  \bibinfo{year}{2016}\natexlab{}.
\newblock \showarticletitle{MIMIC-III, a freely accessible critical care
  database}.
\newblock \bibinfo{journal}{\emph{Scientific data}} \bibinfo{volume}{3},
  \bibinfo{number}{1} (\bibinfo{year}{2016}), \bibinfo{pages}{1--9}.
\newblock


\bibitem[Kinga et~al\mbox{.}(2015)]%
        {kinga2015method}
\bibfield{author}{\bibinfo{person}{D Kinga}, \bibinfo{person}{Jimmy~Ba Adam},
  {et~al\mbox{.}}} \bibinfo{year}{2015}\natexlab{}.
\newblock \showarticletitle{A method for stochastic optimization}. In
  \bibinfo{booktitle}{\emph{International conference on learning
  representations (ICLR)}}, Vol.~\bibinfo{volume}{5}. San Diego, California;,
  \bibinfo{pages}{6}.
\newblock


\bibitem[Le et~al\mbox{.}(2018)]%
        {le2018dual}
\bibfield{author}{\bibinfo{person}{Hung Le}, \bibinfo{person}{Truyen Tran},
  {and} \bibinfo{person}{Svetha Venkatesh}.} \bibinfo{year}{2018}\natexlab{}.
\newblock \showarticletitle{Dual memory neural computer for asynchronous
  two-view sequential learning}. In \bibinfo{booktitle}{\emph{Proceedings of
  the 24th ACM SIGKDD international conference on knowledge discovery \& data
  mining}}. \bibinfo{pages}{1637--1645}.
\newblock


\bibitem[Min and Han(2005)]%
        {min2005recommender}
\bibfield{author}{\bibinfo{person}{Sung-Hwan Min} {and} \bibinfo{person}{Ingoo
  Han}.} \bibinfo{year}{2005}\natexlab{}.
\newblock \showarticletitle{Recommender systems using support vector machines}.
  In \bibinfo{booktitle}{\emph{International Conference on Web Engineering}}.
  Springer, \bibinfo{pages}{387--393}.
\newblock


\bibitem[Ran et~al\mbox{.}(2022)]%
        {ran2022pm}
\bibfield{author}{\bibinfo{person}{Yiding Ran}, \bibinfo{person}{Hengchang Hu},
  {and} \bibinfo{person}{MINYEN KAN}.} \bibinfo{year}{2022}\natexlab{}.
\newblock \showarticletitle{PM K-LightGCN: Optimizing for Accuracy and
  Popularity Match in Course Recommendation}. In
  \bibinfo{booktitle}{\emph{Workshop of Multi-Objective Recommender Systems
  (MORS’22), in conjunction with the 16th ACM Conference on Recommender
  Systems, RecSys}}, Vol.~\bibinfo{volume}{22}. \bibinfo{pages}{2022}.
\newblock


\bibitem[Rodrigues(2019)]%
        {rodrigues2019drug}
\bibfield{author}{\bibinfo{person}{A~David Rodrigues}.}
  \bibinfo{year}{2019}\natexlab{}.
\newblock \bibinfo{booktitle}{\emph{Drug-drug interactions}}.
\newblock \bibinfo{publisher}{CRC Press}.
\newblock


\bibitem[Shang et~al\mbox{.}(2019)]%
        {shang2019gamenet}
\bibfield{author}{\bibinfo{person}{Junyuan Shang}, \bibinfo{person}{Cao Xiao},
  \bibinfo{person}{Tengfei Ma}, \bibinfo{person}{Hongyan Li}, {and}
  \bibinfo{person}{Jimeng Sun}.} \bibinfo{year}{2019}\natexlab{}.
\newblock \showarticletitle{Gamenet: Graph augmented memory networks for
  recommending medication combination}. In
  \bibinfo{booktitle}{\emph{proceedings of the AAAI Conference on Artificial
  Intelligence}}, Vol.~\bibinfo{volume}{33}. \bibinfo{pages}{1126--1133}.
\newblock


\bibitem[Sujan et~al\mbox{.}(2019)]%
        {sujan2019human}
\bibfield{author}{\bibinfo{person}{Mark Sujan}, \bibinfo{person}{Dominic
  Furniss}, \bibinfo{person}{Kath Grundy}, \bibinfo{person}{Howard Grundy},
  \bibinfo{person}{David Nelson}, \bibinfo{person}{Matthew Elliott},
  \bibinfo{person}{Sean White}, \bibinfo{person}{Ibrahim Habli}, {and}
  \bibinfo{person}{Nick Reynolds}.} \bibinfo{year}{2019}\natexlab{}.
\newblock \showarticletitle{Human factors challenges for the safe use of
  artificial intelligence in patient care}.
\newblock \bibinfo{journal}{\emph{BMJ health \& care informatics}}
  \bibinfo{volume}{26}, \bibinfo{number}{1} (\bibinfo{year}{2019}).
\newblock


\bibitem[Sun et~al\mbox{.}(2016)]%
        {sun2016data}
\bibfield{author}{\bibinfo{person}{Leilei Sun}, \bibinfo{person}{Chuanren Liu},
  \bibinfo{person}{Chonghui Guo}, \bibinfo{person}{Hui Xiong}, {and}
  \bibinfo{person}{Yanming Xie}.} \bibinfo{year}{2016}\natexlab{}.
\newblock \showarticletitle{Data-driven automatic treatment regimen development
  and recommendation}. In \bibinfo{booktitle}{\emph{Proceedings of the 22nd ACM
  SIGKDD international conference on knowledge discovery and data mining}}.
  \bibinfo{pages}{1865--1874}.
\newblock


\bibitem[Tatonetti et~al\mbox{.}(2012)]%
        {tatonetti2012data}
\bibfield{author}{\bibinfo{person}{Nicholas~P Tatonetti},
  \bibinfo{person}{Patrick~P Ye}, \bibinfo{person}{Roxana Daneshjou}, {and}
  \bibinfo{person}{Russ~B Altman}.} \bibinfo{year}{2012}\natexlab{}.
\newblock \showarticletitle{Data-driven prediction of drug effects and
  interactions}.
\newblock \bibinfo{journal}{\emph{Science translational medicine}}
  \bibinfo{volume}{4}, \bibinfo{number}{125} (\bibinfo{year}{2012}),
  \bibinfo{pages}{125ra31--125ra31}.
\newblock


\bibitem[Vaswani et~al\mbox{.}(2023)]%
        {vaswani2023attention}
\bibfield{author}{\bibinfo{person}{Ashish Vaswani}, \bibinfo{person}{Noam
  Shazeer}, \bibinfo{person}{Niki Parmar}, \bibinfo{person}{Jakob Uszkoreit},
  \bibinfo{person}{Llion Jones}, \bibinfo{person}{Aidan~N. Gomez},
  \bibinfo{person}{Lukasz Kaiser}, {and} \bibinfo{person}{Illia Polosukhin}.}
  \bibinfo{year}{2023}\natexlab{}.
\newblock \bibinfo{title}{Attention Is All You Need}.
\newblock
\newblock
\showeprint[arxiv]{1706.03762}~[cs.CL]


\bibitem[Wang et~al\mbox{.}(2018)]%
        {wang2018personalized}
\bibfield{author}{\bibinfo{person}{Lu Wang}, \bibinfo{person}{Wei Zhang},
  \bibinfo{person}{Xiaofeng He}, {and} \bibinfo{person}{Hongyuan Zha}.}
  \bibinfo{year}{2018}\natexlab{}.
\newblock \showarticletitle{Personalized prescription for comorbidity}. In
  \bibinfo{booktitle}{\emph{Database Systems for Advanced Applications: 23rd
  International Conference, DASFAA 2018, Gold Coast, QLD, Australia, May 21-24,
  2018, Proceedings, Part II 23}}. Springer, \bibinfo{pages}{3--19}.
\newblock


\bibitem[Wang et~al\mbox{.}(2017)]%
        {wang2017safe}
\bibfield{author}{\bibinfo{person}{Meng Wang}, \bibinfo{person}{Mengyue Liu},
  \bibinfo{person}{Jun Liu}, \bibinfo{person}{Sen Wang},
  \bibinfo{person}{Guodong Long}, {and} \bibinfo{person}{Buyue Qian}.}
  \bibinfo{year}{2017}\natexlab{}.
\newblock \showarticletitle{Safe medicine recommendation via medical knowledge
  graph embedding}.
\newblock \bibinfo{journal}{\emph{arXiv preprint arXiv:1710.05980}}
  (\bibinfo{year}{2017}).
\newblock


\bibitem[Yang et~al\mbox{.}(2021)]%
        {yang2021change}
\bibfield{author}{\bibinfo{person}{Chaoqi Yang}, \bibinfo{person}{Cao Xiao},
  \bibinfo{person}{Lucas Glass}, {and} \bibinfo{person}{Jimeng Sun}.}
  \bibinfo{year}{2021}\natexlab{}.
\newblock \showarticletitle{Change matters: Medication change prediction with
  recurrent residual networks}.
\newblock \bibinfo{journal}{\emph{arXiv preprint arXiv:2105.01876}}
  (\bibinfo{year}{2021}).
\newblock


\bibitem[Yang et~al\mbox{.}(2022)]%
        {yang2022safedrug}
\bibfield{author}{\bibinfo{person}{Chaoqi Yang}, \bibinfo{person}{Cao Xiao},
  \bibinfo{person}{Fenglong Ma}, \bibinfo{person}{Lucas Glass}, {and}
  \bibinfo{person}{Jimeng Sun}.} \bibinfo{year}{2022}\natexlab{}.
\newblock \bibinfo{title}{SafeDrug: Dual Molecular Graph Encoders for
  Recommending Effective and Safe Drug Combinations}.
\newblock
\newblock
\showeprint[arxiv]{2105.02711}~[cs.LG]


\bibitem[Zhang et~al\mbox{.}(2019)]%
        {zhang2019graph}
\bibfield{author}{\bibinfo{person}{Si Zhang}, \bibinfo{person}{Hanghang Tong},
  \bibinfo{person}{Jiejun Xu}, {and} \bibinfo{person}{Ross Maciejewski}.}
  \bibinfo{year}{2019}\natexlab{}.
\newblock \showarticletitle{Graph convolutional networks: a comprehensive
  review}.
\newblock \bibinfo{journal}{\emph{Computational Social Networks}}
  \bibinfo{volume}{6}, \bibinfo{number}{1} (\bibinfo{year}{2019}),
  \bibinfo{pages}{1--23}.
\newblock


\bibitem[Zhang et~al\mbox{.}(2017)]%
        {zhang2017leap}
\bibfield{author}{\bibinfo{person}{Yutao Zhang}, \bibinfo{person}{Robert Chen},
  \bibinfo{person}{Jie Tang}, \bibinfo{person}{Walter~F Stewart}, {and}
  \bibinfo{person}{Jimeng Sun}.} \bibinfo{year}{2017}\natexlab{}.
\newblock \showarticletitle{LEAP: learning to prescribe effective and safe
  treatment combinations for multimorbidity}. In
  \bibinfo{booktitle}{\emph{proceedings of the 23rd ACM SIGKDD international
  conference on knowledge Discovery and data Mining}}.
  \bibinfo{pages}{1315--1324}.
\newblock


\bibitem[Zhang et~al\mbox{.}(2023)]%
        {zhang2023knowledge}
\bibfield{author}{\bibinfo{person}{Yingying Zhang}, \bibinfo{person}{Xian Wu},
  \bibinfo{person}{Quan Fang}, \bibinfo{person}{Shengsheng Qian}, {and}
  \bibinfo{person}{Changsheng Xu}.} \bibinfo{year}{2023}\natexlab{}.
\newblock \showarticletitle{Knowledge-enhanced attributed multi-task learning
  for medicine recommendation}.
\newblock \bibinfo{journal}{\emph{ACM Transactions on Information Systems}}
  \bibinfo{volume}{41}, \bibinfo{number}{1} (\bibinfo{year}{2023}),
  \bibinfo{pages}{1--24}.
\newblock


\end{thebibliography}

\appendix

\end{document}